\documentclass[a4paper]{article}
\usepackage{bm}
\usepackage{multirow}
\usepackage{INTERSPEECH2016}
\usepackage{graphicx}
\usepackage{amssymb,amsmath,bm}
\usepackage{textcomp}
\usepackage{subfig}
\usepackage{multirow}
\usepackage{tabularx}

\ninept

\title{Query-based Attention CNN for Text Similarity Map}

\makeatletter
\def\name#1{\gdef\@name{#1\\}}
\makeatother \name{{\em Tzu-Chien Liu*, Yu-Hsueh Wu*, Hung-Yi Lee} \\
	\footnotesize *These authors contributed to the work equally and should be regarded as co-first authors}

\address{Graduate Institute of Communication Engineering, National Taiwan University \\
  {\footnotesize \tt \{b01901153, b01901062, hungyilee\}@ntu.edu.tw}
}

\begin{document}

\maketitle
  %
\begin{abstract}

  In this paper, we introduce Query-based Attention CNN(QACNN) for Text Similarity Map, an end-to-end neural network for question answering. This network is composed of compare mechanism, two-staged CNN architecture with attention mechanism, and a prediction layer. First, the compare mechanism compares between the given passage, query, and multiple answer choices to build similarity maps. Then, the two-staged CNN architecture extracts features through word-level and sentence-level. At the same time, attention mechanism helps CNN focus more on the important part of the passage based on the query information. Finally, the prediction layer find out the most possible answer choice. We conduct this model on the MovieQA~\cite{MovieQA} dataset using Plot Synopses only, and achieve 79.99\% accuracy which is the state of the art on the dataset.

  \end{abstract}


  \section{Introduction}
  	  Many machine learning models in question answering tasks often involve matching mechanism. For example, in factoid question answering such as SQuAD~\cite{rajpurkar2016squad}, one needs to match between query and corpus in order to find out the most possible fragment as answer. In multiple choice question answering, such as MC Test~\cite{richardson2013mctest}, matching mechanism can also help make the correct decision. 
    
    The easiest way of matching is to calculate the cosine similarity between two vectors. It is generally done by two step: First, encode text into word vectors, sentence vectors or paragraph vectors. Second, simply calculate the cosine similarity between target vectors. This method performs well when applied to word-level matching. However, as for matching between sentences or paragraphs, a single vector is not sufficient to encode all the important information. In order to solve this problem, Wang and Jiang proposed a “compare-aggregate”~\cite{wang2016compare} framework that performs word-level matching using multiple techniques followed by aggregation with convolutional neural network. In their work, they show that compare-aggregate framework can effectively match two sequences through a wide range. 
    
    Although "compare-aggregate" matching mechanism performs well on multiple question answering tasks, it has two deficiencies. First, it tends to aggregate passively through the sequence rather than take the importance of each element into account. That is, "compare aggregate" model considers all the sequential contents equally.  Second, "compare aggregate" can only take few neighboring elements into account at the same time because of the limitation of CNN kernel size.

In this paper, we propose Query-based Attention CNN (QACNN) to deal with the deficiencies above. First, we add query-based attention mechanism into original "compare aggregate" model. Moreover, We re-design the aggregation mechanism in "compare aggregate" to a two-staged CNN architecture which comprises word-level aggregation and sentence-level aggregation. In this way, QACNN can efficiently extract features cross sentences.
  
  Our model consists of three components: 1) The similarity mapping layer which converts the input passage, query and choice into feature representation and perform a similarity operation to each other. 2) The attention-based CNN matching network composed of a two-staged CNN focusing on word-level and sentence-level matching respectively. 3) The prediction layer which makes the final decision.
  
  The main contributions of this work are three-fold. First, we introduce a two-staged CNN architecture which integrates information from word-level to sentence-level, and then from sentence-level to passage-level. Second, we introduce attention mechanism into this net.  We use specially designed CNN structure and attention mechanism to recognize the pattern of similarity map and eventually identify specific syntactic structure of queries.  By transforming passage-query feature into attention maps and applying it to passage-choice matching result, we reasonably give weight to every word in the passage.  
    Lastly, our model reaches 79.99\% accuracy on the MovieQA dataset which yields top 1 result on this dataset.

\begin{figure}[ht]
\begin{center}
\includegraphics[scale=0.8]{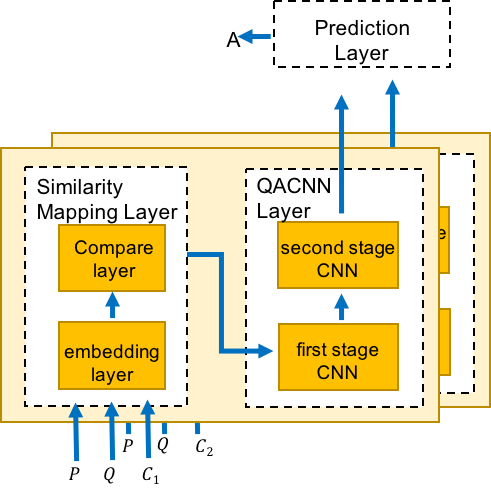}
        \caption{{\it QACNN overview, P denotes paragraph, Q denotes query, C denotes one of choices}}
        \label{fig:ACM_pipeline}
\end{center}
\end{figure}

\section{QACNN}
In this question answering task, a reading passage , a query and several answer choices are given. \textbf{P} denotes the passage, \textbf{Q} denotes query and \textbf{C} denotes one of the multiple choices. The target of the model is to choose a correct answer \textbf{A} from multiple choices based on informations of  \textbf{P} and \textbf{Q}.


Fig.\ref{fig:ACM_pipeline} is the pipeline overview of QACNN.  First, we use embedding layer to transform \textbf{P}, \textbf{Q}, and \textbf{C} into word embedding. 
Then the compare layer generates passage-query similarity map $\bm{PQ}$ and passage-choice similarity map $\bm{PC}$. The following part is the main component of QACNN. It consists of two-staged CNN architecture. The first stage projects word-level feature into sentence-level, and the second stage projects sentence-level feature into passage-level. Moreover, we apply query-based attention mechanism to each stage on the basis of $\bm{PQ}$ feature at word level and sentence level respectively. After QACNN Layer, we obtain each choice answer feature. Finally, a prediction layer collects output information from every choice feature and returns the most possible answer.

\subsection{Similarity Mapping Layer}
Similarity Mapping Layer is composed of two part: embedding layer and compare layer. Given a passage \textbf{P} with $N$ sentences, a query \textbf{Q}, and  a choice \textbf{C}, the embedding layer transforms every words in \textbf{P}, \textbf{Q} and \textbf{C} into word embedding\footnote{Both query and choice are considered as a sentence.}:
\begin{equation}
	\begin{split}
		&\bm{P} = \{p_n^i\}_{i=1,n=1}^{I,N}\\
		&\bm{Q} = \{q^j\}_{j=1}^J\\
		&\bm{C} = \{c^k\}_{k=1}^K
	\end{split}
\end{equation}
$I$ is the length of a sentence in passage\footnote{By padding, all the sentences in all the passages have the same length.}, and $J$ and $K$ are the length of query and length of one single choice respectively\footnote{We also make all the queries have the same length $J$, and all the choices have the same length $K$ by padding.}. 
$p_n^i$, $q^j$ and $c^k$  are word embeddings.
Word embedding can be obtained by any type of embedding technique, such as recurrent neural network\cite{mikolov2010recurrent}, Sequence-to-sequence model\cite{sutskever2014sequence}, Word2vec\cite{goldberg2014word2vec}, etc. In our work, we simply use pre-trained GloVe word vectors\cite{pennington2014glove} as the embedding without any further modification or training. 

\begin{figure}[t]
        \centering
        \includegraphics[width=\linewidth]{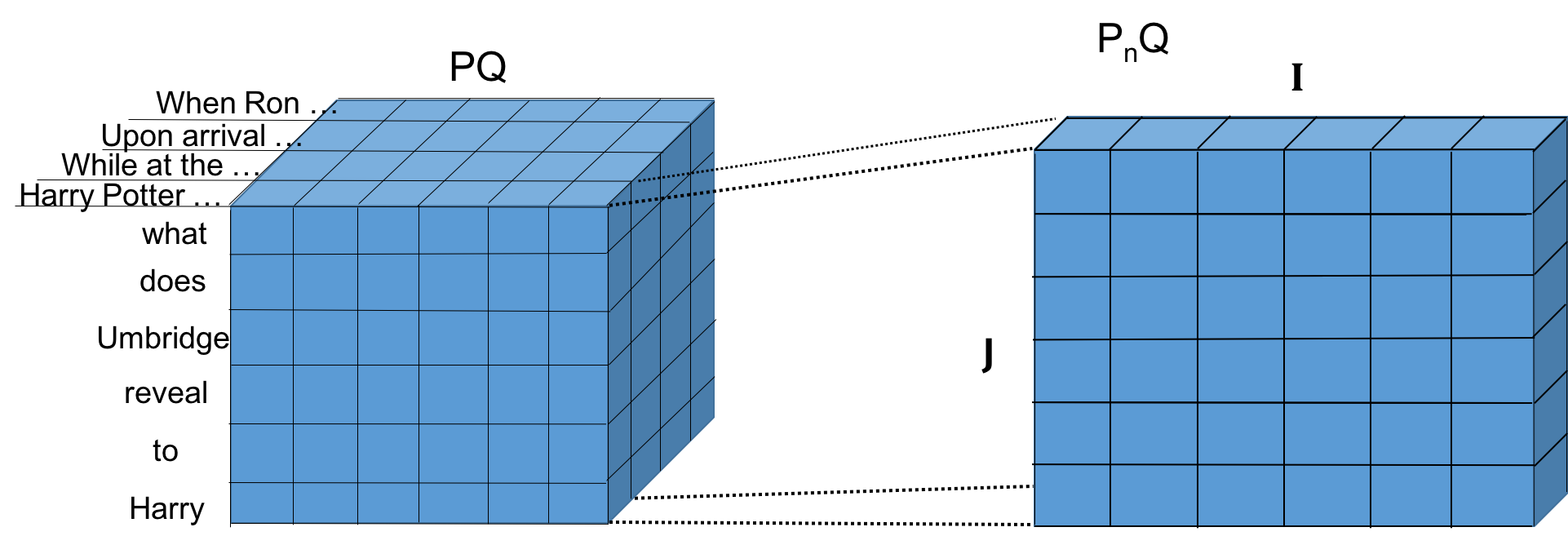}
        \caption{{\it Similarity map between paragraph  \textbf{P} and query \textbf{Q}. 
        \textbf{I} denotes the length of each sentence $\bm{P_n}$, \textbf{J} denotes the length of query \textbf{Q}}}
        \label{fig:similarity_map.png}
\end{figure}
\begin{figure}[t]
        \centering
        \includegraphics[width=\linewidth]{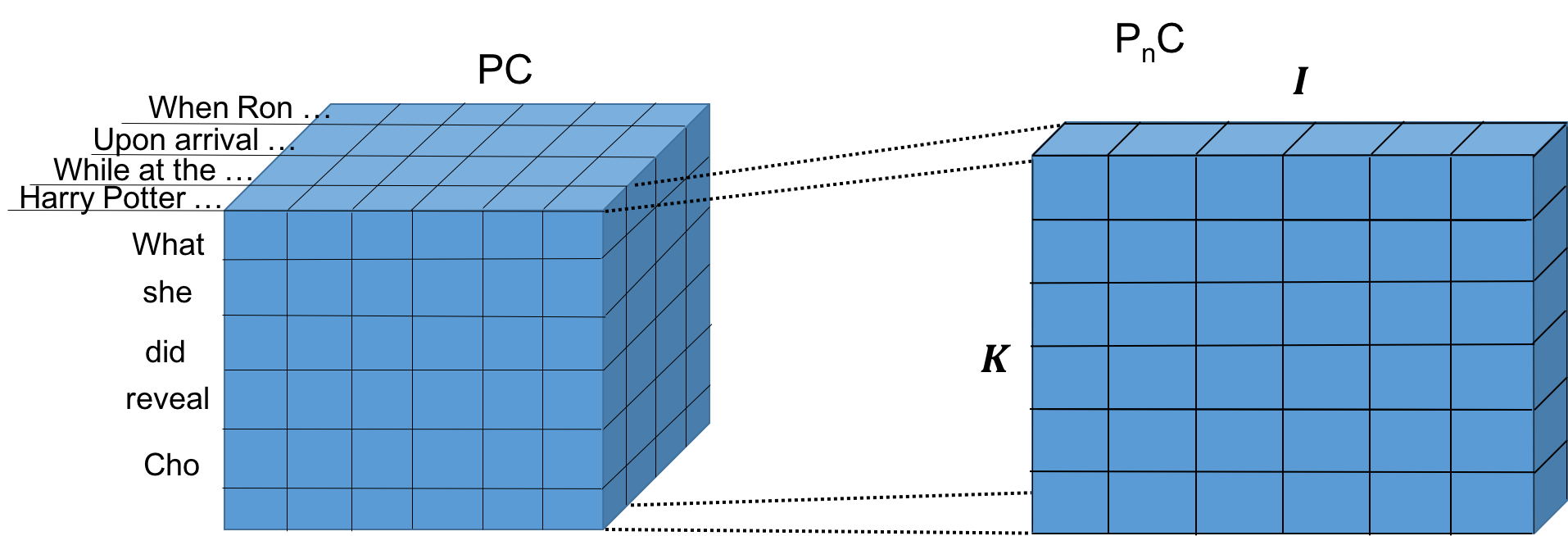}
        \caption{{\it Similarity map between paragraph \textbf{P} and choice \textbf{C}. 
        \textbf{I} denotes the length of each sentence $\bm{P_n}$, K denotes the length of choice \textbf{C}}}
        \label{fig:similarity_map_C.png}
\end{figure}
After word embedding step, We want to acquire similarity map which tells us location relationship between passage and query, passage and choices. We use compare layer to compare each passage sentence $\bm{P_n}$ to $\bm{Q}$ and $\bm{C}$ at word level separately as Fig.\ref{fig:similarity_map.png} and Fig.\ref{fig:similarity_map_C.png} show. 
\begin{equation}
	\begin{split}
		&\bm{P_nQ} = \{cos(p_n^i,q^j)\}_{i=1,j=1}^{I,J}\\
        &\bm{P_nC} = \{cos(p_n^i,c^k)\}_{i=1,k=1}^{I,K}
	\end{split}
 \end{equation}
That is, we compare each word in sentences of passage to each word in query and choice. We use cosine similarity as the comparing method here. 
This step creates two similarity map, passage-query similarity map $\bm{PQ = [P_1Q,P_2Q,...,P_NQ]\in\mathbb{R}^{N\times J\times I}}$ and passage-choice similarity map $\bm{PC = [P_1C,P_2C,...,P_NC]\in\mathbb{R}^{N\times K\times I}}$.

\begin{figure}[t]
        \centering
        \includegraphics[width=\linewidth]{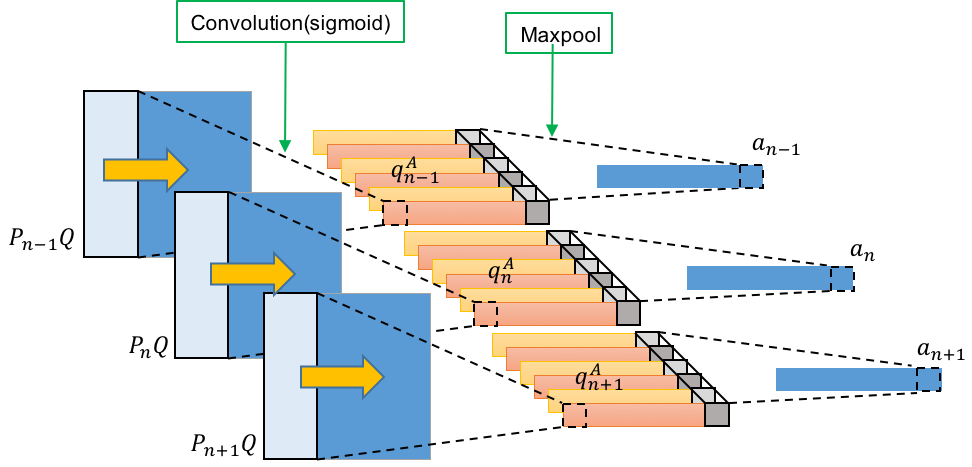}
        \caption{{\it First stage CNN attention part.}}
        \label{fig:CNN1-1.png}
\end{figure} 

\begin{figure}[t]
        \centering
        \includegraphics[width=\linewidth]{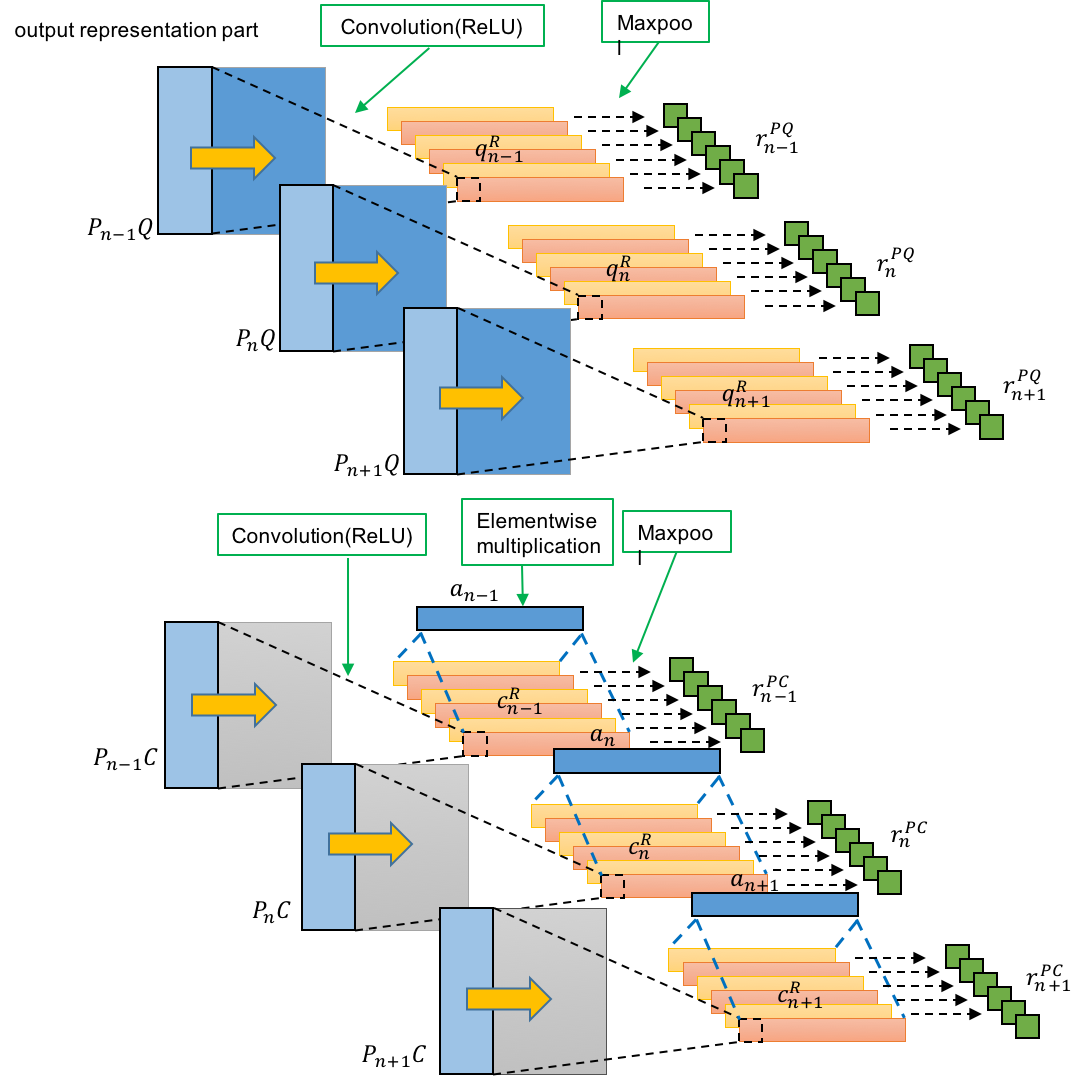}
        \caption{{\it First stage CNN representation part. $\bm{a_{n}}$ is the word-level attention from First stage CNN attention.}}
        \label{fig:CNN1-2.png}
\end{figure} 

\subsection{QACNN Layer}
\label{sec:acmlayer} 
We propose an attention convolutional matching layer to integrate two similarity maps given above. That is, QACNN Layer is used to learn the  location relationship pattern. It contains a two-staged CNN combined with query-based attention mechanism. Each stage comprises two major part: attention map and output representation. 

\subsubsection{Attention Map of First Stage}
\label{sec:attention_map_first_stage} 
Fig.\ref{fig:CNN1-1.png} shows the architecture of the attention map in first stage CNN.  We choose $n^{th}$ sentence slice $\bm{P_nQ\in\mathbb{R}^{J\times I}}$ in $\bm{PQ}$, and apply CNN to it using the convolution kernel $\bm{W_1^{A}\in\mathbb{R}^{J\times l\times d}}$, where superscript $\bm{A}$ denotes attention map, subscript $\bm{1}$ denotes the first stage CNN. Symbol $d$ and $l$ represent width of kernel and  number of kernel respectively.  The generated feature $\bm{q_n^A\in\mathbb{R}^{l\times (I-d+1)}}$ is as follow:
\begin{equation} \label{eq:2}
	\begin{split}
		&\bm{q_n^A = sigmoid(W_1^{A}\ast P_nQ+b_1^{A})}
	\end{split}
\end{equation}
where $\bm{b_1^{A}\in\mathbb{R}^l}$ is the bias. With $\bm{W_1^{A}}$ covering whole query and several words in the passage, convolution kernels would learn the query syntactic structure and give weight to each passage's location. That's why we use sigmoid function as activation function in this stage. Furthermore, we perform maxpooling to $\bm{q_n^A}$ perpendicularly in order to find the largest weight between different kernels in the same location, using maxpool kernel shaped $l$, and then generate word-level attention map $\bm{a_{n}\in\mathbb{R}^{I-d+1}}$ for each sentence.

\subsubsection{Output Representation of First Stage}
\label{sec:representation_first_stage}
In this stage, we want to acquire passage's sentence features based on the query and the choice respectively. We apply CNN to $\bm{P_nC}$ to aggregate pattern of location relationship and acquire choice-based sentence features. Also, we apply CNN to $\bm{P_nQ}$ to acquire query-based sentence features. CNN architecture of output representation part Fig.\ref{fig:CNN1-2.png} is similar to which of attention map part, but we use different kernels $\bm{W_1^{R}\in\mathbb{R}^{l\times K\times d}}$ and different bias $\bm{b_1^{R}\in\mathbb{R}^l}$:
\begin{equation} \label{eq:3}
	\begin{split}
		&\bm{q_n^R = ReLU(W_1^{R}\ast P_nQ+b_1^{R})}\\
        &\bm{c_n^R = ReLU(W_1^{R}\ast P_nC+b_1^{R})}
	\end{split}
\end{equation}
where the superscript $\bm{R}$ denotes output representation.\footnote{different choices share same $\bm{W_1^{R}}$ and $\bm{b_1^{R}}$.}
We apply $\bm{W_1^{R}}$ on $\bm{P_nC}$ and $\bm{P_nQ}$ then finally generate $\bm{q_n^R}$ and $\bm{c_n^R\in\mathbb{R}^{l\times (I-d+1)}}$ using eq.4.  We then multiply $\bm{c_n^R}$ by the word-level attention map $\bm{a_{n}}$ which comes from stage 2.2.1 element-wise through the first dimension. At last, we maxpool $\bm{q_n^R}$ and $\bm{c_n^R}$ horizontally with kernel shape $(I-d+1)$ to get the query-based sentence features $\bm{r_n^{PQ}}$ and choice-based sentence features $\bm{r_n^{PC}\in\mathbb{R}^l}$.

\subsubsection{Attention Map of Second Stage}   
\label{sec:attention_map_second_stage} 
\begin{figure}[t]
        \centering
        \includegraphics[width=\linewidth]{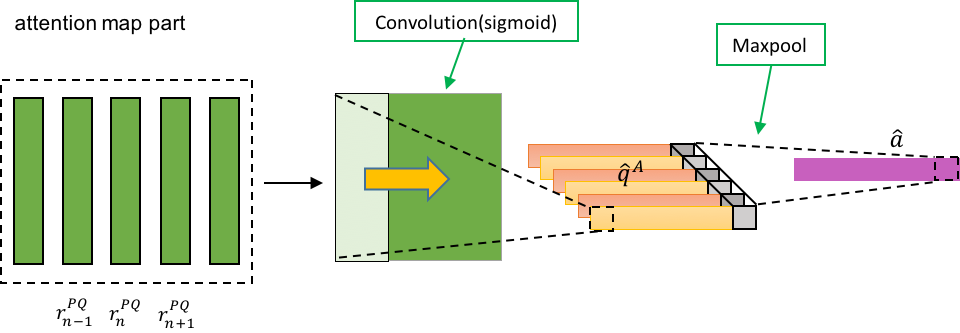}
        \caption{{\it Second stage CNN attention part.}}
        \label{fig:CNN2-1.png}
\end{figure}

Fig.\ref{fig:CNN2-1.png} is the architecture of attention map in the second stage CNN. Based on first stage query-based sentence features from section~\ref{sec:representation_first_stage}, we want to acquire sentence-level attention map. The input of this stage is $\bm{r^{PQ}=[r_1^{PQ},r_2^{PQ},...,r_N^{PQ}]}$, which will be further refined by CNN with kernel $\bm{W_2^{A}\in\mathbb{R}^{l\times d\times l}}$ and generates intermediate features 
$\bm{\hat{q}^{A}\in\mathbb{R}^{l\times (N-d+1)}}$.
\begin{equation} \label{eq:4}
	\begin{split}
		&\bm{\hat{q}^{A} = sigmoid(W_2^{A}\ast r^{PQ}+b_2^{A})}
	\end{split}
\end{equation}
Then, same as attention map of first stage, we maxpool $\bm{\hat{q}^{A}}$ with kernel shaped $l$, and obtain sentence-level attention map $\bm{\hat{a}\in\mathbb{R}^{N-d+1}}$. 

\subsubsection{Output Representation of Second Stage}
\begin{figure}[t]
        \centering
        \includegraphics[width=\linewidth]{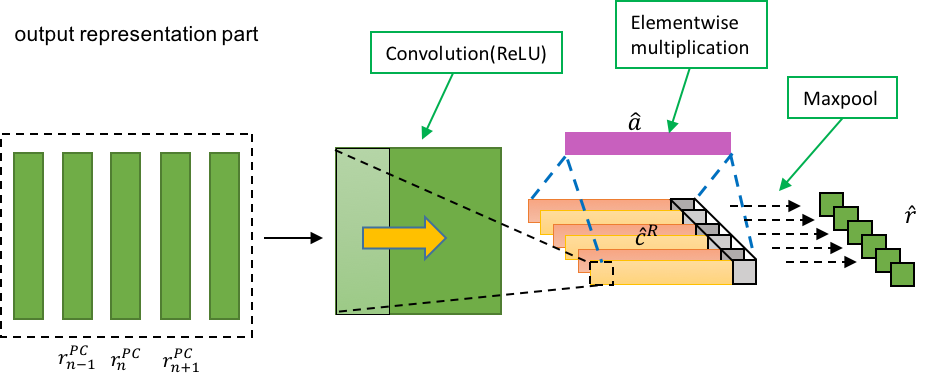}
        \caption{{\it Second stage CNN representation part. $\bm{\hat{a}}$ is the sentence-level attention from Second stage CNN attention.}}
        \label{fig:CNN2-2.png}
\end{figure}

Output representation part of the second stage  in Fig.\ref{fig:CNN2-2.png} has two input, sentence-level attention map $\bm{\hat{a}}$ and sentence-level features $\bm{r^{PC}=[r_1^{PC},r_2^{PC},...,r_N^{PC}]}$. The equations here are similar to those previously mentioned. As follow: 
\begin{equation} \label{eq:5}
	\begin{split}
		&\bm{\hat{c}^{R} = ReLU(W_2^{R}\ast r^{PC}+b_2^{R})}\\
        &\bm{\hat{r} = \{max(\hat{c}_t^{R}\cdot\hat{a})}\}_{t=1}^l
	\end{split}
\end{equation}
where $\bm{W_2^{R}\in\mathbb{R}^{l\times l\times d}}$, $\bm{b_2^{R}\in\mathbb{R}^l}$, and $\bm{\hat{c}^{R}\in\mathbb{R}^{l\times (N-d+1)}}$. Output representation of certain choice $\bm{\hat{r}\in\mathbb{R}^l}$ is the final output of QACNN Layer.

\subsection{Prediction Layer}

Prediction Layer is the final part of QACNN. We use  $\bm{\hat{r}_m\in\textbf{R}^l}$ to represent the final output representation of the $\bm{m^{th}}$ choice . In order to find out the most correct choice, we simply pass $\bm{\hat{r}_m}$ to two fully-connected layer and compute probability for each choice using softmax as follows:
\begin{equation} \label{eq:7}
	\begin{split}
    	&\bm{R = \{\hat{r}_m\}_{m=1}^{M}}
	\end{split}
\end{equation}
\begin{equation} \label{eq:8}
	\begin{split}
    	&\bm {p(m|R) = softmax(W^{o}(tanh(W^{p}R+b^{p}))+b^{o} )}
	\end{split}
\end{equation}
where $\bm{W^{p}\in\mathbb{R}^{l\times l}}$,$\bm{b^{p}\in\mathbb{R}^{l}}$, $\bm{W^{o}\in\mathbb{R}^{l}}$, and $\bm{b^{o}\in\mathbb{R}}$.

\section{Experiments}

\subsection{Implementation details}
In the preprocessing step, we used pre-trained GloVe vectors for word embeddings, and they would not be updated during training; We padded sentence number in each passage to 101, all word number in each sentence to 100. Word number of queries and choices were padded to 50.  For all kernels of CNN, $\bm{W_1^{A}},\bm{W_2^{A}},\bm{W_1^{R}},\bm{W_2^{R}}$, each of which has three different kernel width $\bm{d}=\{1,3,5\}$; each of them has same kernel number $\bm{l}=128$. We utilized dropout in each CNN layer with dropout rate 0.8. We used Adam~\cite{kingma2014adam} optimizer to optimize our model with initial learning rate 0.001. 

\subsection{Experimental Result}

\subsubsection{MovieQA Result}

We mainly focus on the MovieQA dataset to train and evaluate our model. 
MovieQA dataset aims to evaluate automatic story comprehension from both video and text. The data set consists of almost 15,000 multiple choice question answers. Diverse information in this dataset like plots, scripts, sub-title and video captions can be used to infer answers. In our task, only plot informations are used. This challenging dataset is suitable to evaluate QACNN because movie plots are longer than normal reading comprehension task.
Each question comes with a set of five highly plausible choices, only one of which is correct; 
In the MovieQA benchmark, there are 1958 QA pairs in the val set and 3138 QA pairs in the test set. 

We used ensemble model in this dataset. The ensemble model consists of eight training runs models with identical structure and hyper-parameter. 
In the val set, we achieve 77.6\% accuracy with single model and 79.0\% accuracy with ensemble model. In the test set, as the Table 1. shows, our model achieves 79.99 \% accuracy with ensemble model and is the state of the art.

\begin{table}[hb]
\centering
\caption{MovieQA result~\cite{MovieQA},~\cite{wang2016compare}}
\label{my-label}
\begin{tabular}{|l|l|l|}
\hline
Models & \multicolumn{1}{c|}{dev set } & \multicolumn{1}{c|}{test set} \\ \hline
Cosine Word2Vec & \multicolumn{1}{c|}{46.4} & \multicolumn{1}{c|}{45.63} \\ \hline
Cosine TFIDF & \multicolumn{1}{c|}{47.6} & \multicolumn{1}{c|}{47.36} \\ \hline
SSCB TFIDF & \multicolumn{1}{c|}{48.5} & \multicolumn{1}{c|}{-} \\ \hline
Compare Aggregate & \multicolumn{1}{c|}{72.1} & \multicolumn{1}{c|}{72.9} \\ \hline
QACNN & \multicolumn{1}{c|}{77.6} & \multicolumn{1}{c|}{75.84}\\ \hline
Convnet Fusion & \multicolumn{1}{c|}{-}  & \multicolumn{1}{c|}{77.63} \\ \hline
{\bf QACNN(ensemble)} & \multicolumn{1}{c|}{{\bf79.0}} & \multicolumn{1}{c|}{{\bf79.99}} \\ \hline
\end{tabular}
\end{table}

\begin{figure*}[h]
        \centering
        \includegraphics[width=\linewidth]{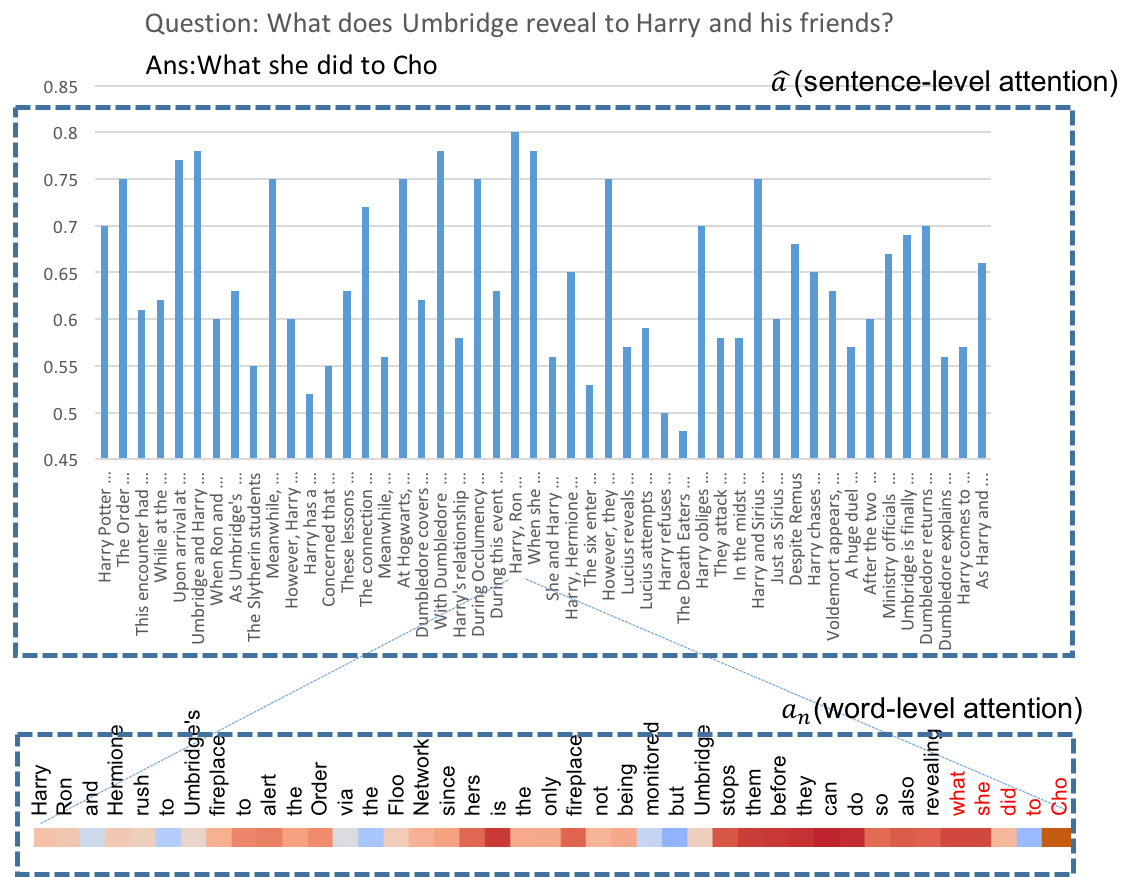}
        \caption{{\it Visualization of the attention map. $\bm{\hat{a}}$ is the sentence-level attention from Second stage CNN attention. $\bm{a_{n}}$ is the word-level attention from First stage CNN attention.}}
        \label{fig:attention_map.png}
\end{figure*}

\subsubsection{MCTest Result}
We also applied our model to MCTest dataset which requires machines to answer multiple-choice reading comprehension questions about fictional stories. The original paper describes that a baseline method uses a combination of a sliding window score and a distance based . They achieve 66.7\% and 56.7\% on MC500 and MC160 separately. Because of the restricted training set and development set, we trained our model on MovieQA training set and applied the result to test MCTest dataset.
On MCTest dataset, we still outperform baseline and achieve 68.1\% accuracy on MC160 and 61.5\% accuracy on MC500.

\subsection{Discussion}

QACNN is a powerful network focusing on multiple choice QA task. It matches between passage and choices based on query information. One of the most important idea in QACNN is two-staged attention map. The first attention map is at word level, representing the importance of each word in paragraph to a certain question; the second attention map, however is at sentence level, representing the importance of each sentence in paragraph to a certain question. 

In this section, we designed several experiments to test how two-stage mechanism and attention maps impact on our model. 
\subsubsection{Two-stage Effect Experiment}
In this experiment,we focused on the difference between one-stage QACNN and two-stage QACNN. For one-stage QACNN, we didn't split an entire passage into sentences. That is, the shape of passage-query similarity map $\bm{PQ}$ and passage-choice similarity map $\bm{PC}$ are 2D rather than 3D. We convolved them directly on word-level and output passage feature without second-stage involved. The result is shown on table~\ref{experiment-label}. The result shows that the modified one staged QACNN reaches 66.8\% accuracy on validation set, which is ten percent lower than 78.1\%, the original QACNN accuracy on validation set.

\subsubsection{Attention Effect Experiment}
In this experiment, our target is to validate the effect of query-based attention in QACNN. We modified three different structures from original QACNN Layer below:

1) For the first one, we modified QACNN Layer in section \ref{sec:acmlayer} and removed both sentence-level attention map and word-level attention map part from it. However, this modified model would have a deficiency of query information. Therefore, we concatenated the final output representation of $\bm{PQ}$ and $\bm{PC}$ together before prediction layer. The experiment result is shown on the Table~\ref{experiment-label}. The result is almost ten percent less than the original one. 2) For the second one, we only removed sentence-level attention in section \ref{sec:attention_map_second_stage} from QACNN Layer and kept word-level attention in the model. 3) For the last one, instead of removing sentence-level attention, we removed word-level attention from QACNN Layer.

The result is shown in Table~\ref{experiment-label}. We can see that QACNN(with only word-level attention) performs better than QACNN(without attention); QACNN(with only sentence-level attention) performs better than QACNN(with only word-level attention); And original QACNN which contains both word-level and sentence-level attention does the best job among all. Thus, not only word-level attention but also sentence-level attention can contribute to the performance of QACNN. However, sentence-level attention seems to play a more important role.

\begin{table}[hb]
\centering
\caption{Experiment result}
\label{experiment-label}
\begin{tabular}{|l|l|l|}
\hline
Models & \multicolumn{1}{c|}{dev set } & \multicolumn{1}{c|}{test set} \\ \hline
One stage QACNN& \multicolumn{1}{c|}{66.8} & \multicolumn{1}{c|}{-} \\ \hline
QACNN(no attention) & \multicolumn{1}{c|}{69.6} & \multicolumn{1}{c|}{-} \\ \hline
QACNN(only word-level attention) & \multicolumn{1}{c|}{72.5} & \multicolumn{1}{c|}{-} \\ \hline
QACNN(only sentence-level attention) & \multicolumn{1}{c|}{75.1} & \multicolumn{1}{c|}{-} \\ \hline
QACNN(single) & \multicolumn{1}{c|}{77.6} & \multicolumn{1}{c|}{75.84}\\ \hline
{\bf QACNN(ensemble)} & \multicolumn{1}{c|}{{\bf79.0}} & \multicolumn{1}{c|}{{\bf79.99}} \\ \hline
\end{tabular}
\end{table}

\subsubsection{Attention Effect Discussion}
Figure.\ref{fig:attention_map.png} is the visualization of two attention maps and their corresponding question. The upper half of Figure.\ref{fig:attention_map.png} is the attention map at sentence level. We picked the sentence with the largest attention value as target sentence and examine it. Thus, we could get the lower half of Figure.\ref{fig:attention_map.png}, which shows the attention map at word level on the target sentence. We used a question from movie, Harry Potter, as an example. The result shows that the sentence with the largest attention value is exactly where the correct answer comes from. It turns out that sentence-level attention map can successfully find out which sentence contains the information of correct answer. As for word-level attention map, we can easily see that the attention map focus mainly on the end of target sentence, which is obviously more important for answering this question.

\section{Conclusion}

In this paper, we present an efficient matching mechanism on multiple choice question answering task. We introduce two-staged CNN to match passage and choice on word level and sentence level. In addition, we use query-based CNN attention to enhance matching effect. 

The power of the model is verified on MovieQA dataset, which yielded the state of the art result on the dataset.
In the future, we are now working on training our model based on our own trained embedding with TF-IDF ~\cite{ramos2003using} weighting.  
Furthermore, we would like to test our model on open-answer task like SQuaD by seeing the whole corpus as an ``answer pool" and solve it like multiple choice question. 
 
\newpage
\eightpt

\bibliographystyle{IEEEtran}
\bibliography{mybib}

\end{document}